\documentclass[letterpaper]{article} 
\usepackage{aaai24}  
\usepackage{times}  
\usepackage{helvet}  
\usepackage{courier}  
\usepackage[hyphens]{url}  
\usepackage{graphicx} 
\usepackage{natbib}  
\usepackage{caption} 
\usepackage{algorithm}
\usepackage{algorithmic}

\usepackage{amssymb}
\usepackage{amsmath}
\usepackage[center]{subfigure}

\usepackage{multirow}
\usepackage{mathtools}
\usepackage{stmaryrd}
\usepackage{gensymb}
\usepackage{newfloat}
\usepackage{listings}
\DeclareCaptionStyle{ruled}{labelfont=normalfont,labelsep=colon,strut=off} 
\floatstyle{ruled}
\newfloat{listing}{tb}{lst}{}
\floatname{listing}{Listing}
\newcommand{\mynetwork}{SPEAL}

\newcommand{\mynetworklongbf}{\textbf{S}keletal \textbf{P}rior \textbf{E}mbedded \textbf{A}ttention \textbf{L}earning}

\newcommand{\sem}{SEM}
\newcommand{\semfullbf}{\textbf{S}keleton \textbf{E}xtraction \textbf{M}odule}
\newcommand{\semfullcap}{Skeleton Extraction Module}

\newcommand{\sagtr}{SAGTR}
\newcommand{\sagtrfullbf}{\textbf{S}keleton-\textbf{A}ware \textbf{G}eo\textbf{TR}ansformer}
\newcommand{\sagtrfullcap}{Skeleton-Aware GeoTransformer}

\newcommand{\cds}{CDS}
\newcommand{\cdsfullbf}{\textbf{C}orrespondence \textbf{D}ual-\textbf{S}ampler}
\newcommand{\cdsfullcap}{Correspondence Dual-Sampler}

\newcommand{\ie}{\textit{i}.\textit{e}., }

\newcommand{\germanyforest}{\mbox{GermanyForest3D}}
\newcommand{\kittics}{\mbox{KITTI CrossSource}}

\usepackage{color,xcolor}
\newcommand{\qs}[1]{\textcolor{blue}{{#1}}}
\newcommand\customparagraph[1]{\noindent\textbf{#1.}}

\newcommand{\xkznew}[1]{\textcolor{red}{{#1}}}
\newcommand{\xkzdel}[1]{}

\renewcommand{\textcolor}[2]{#2}

\title{\mynetwork: Skeletal Prior Embedded Attention Learning for Cross-Source Point Cloud Registration}
\author {
    Kezheng Xiong\textsuperscript{\rm 1},  
    Maoji Zheng\textsuperscript{\rm 1}, 
    Qingshan Xu\textsuperscript{\rm 2}, 
    Chenglu Wen\textsuperscript{\rm 1}\footnote{Corresponding author.},  
    Siqi Shen\textsuperscript{\rm 1}\footnotemark[1], 
    Cheng Wang\textsuperscript{\rm 1}
}
\affiliations {
    \textsuperscript{\rm 1} Xiamen University
    \textsuperscript{\rm 2} Nanyang Technological University\\
    \{xiongkezheng, zhengmaoji\}@stu.xmu.edu.cn,
    qingshan.xu@ntu.edu.sg,\
    \{clwen, siqishen, cwang\}@xmu.edu.cn,
}

\usepackage{bibentry}

\begin{document}

\maketitle

\begin{abstract}

Point cloud registration, a fundamental task in 3D computer vision, has remained largely unexplored in cross-source point clouds and unstructured scenes.
The primary challenges arise from noise, outliers, and variations in scale and density.
However, neglected geometric natures of point clouds restrict the performance of current methods.
In this paper, we propose a novel \qs{method}, termed \mynetwork{}, to leverage skeletal representations for effective learning of intrinsic topologies of point clouds, facilitating robust capture of geometric intricacy.
Specifically, we design the \semfullcap{} to extract skeleton points and skeletal features in an unsupervised manner, which is inherently robust to noise and density variances.
Then, we propose the \sagtrfullcap{} to encode high-level skeleton-aware features.
It explicitly captures the topological natures and inter-point-cloud skeletal correlations with the noise-robust and density-invariant skeletal representations. 
Next, we introduce the \cdsfullcap{} to facilitate correspondences by augmenting the correspondence set with skeletal correspondences.
Furthermore, we construct a challenging novel cross-source point cloud dataset named \kittics{} for benchmarking cross-source point cloud registration methods.
Extensive quantitative and qualitative experiments are conducted to demonstrate our approach's superiority and robustness on both cross-source and same-source datasets.
To the best of our knowledge, our approach is the first to facilitate point cloud registration with skeletal geometric priors. 

\end{abstract}

\section{Introduction}


\begin{figure}[ht]
  \centering
  \includegraphics[width=1.0\linewidth]{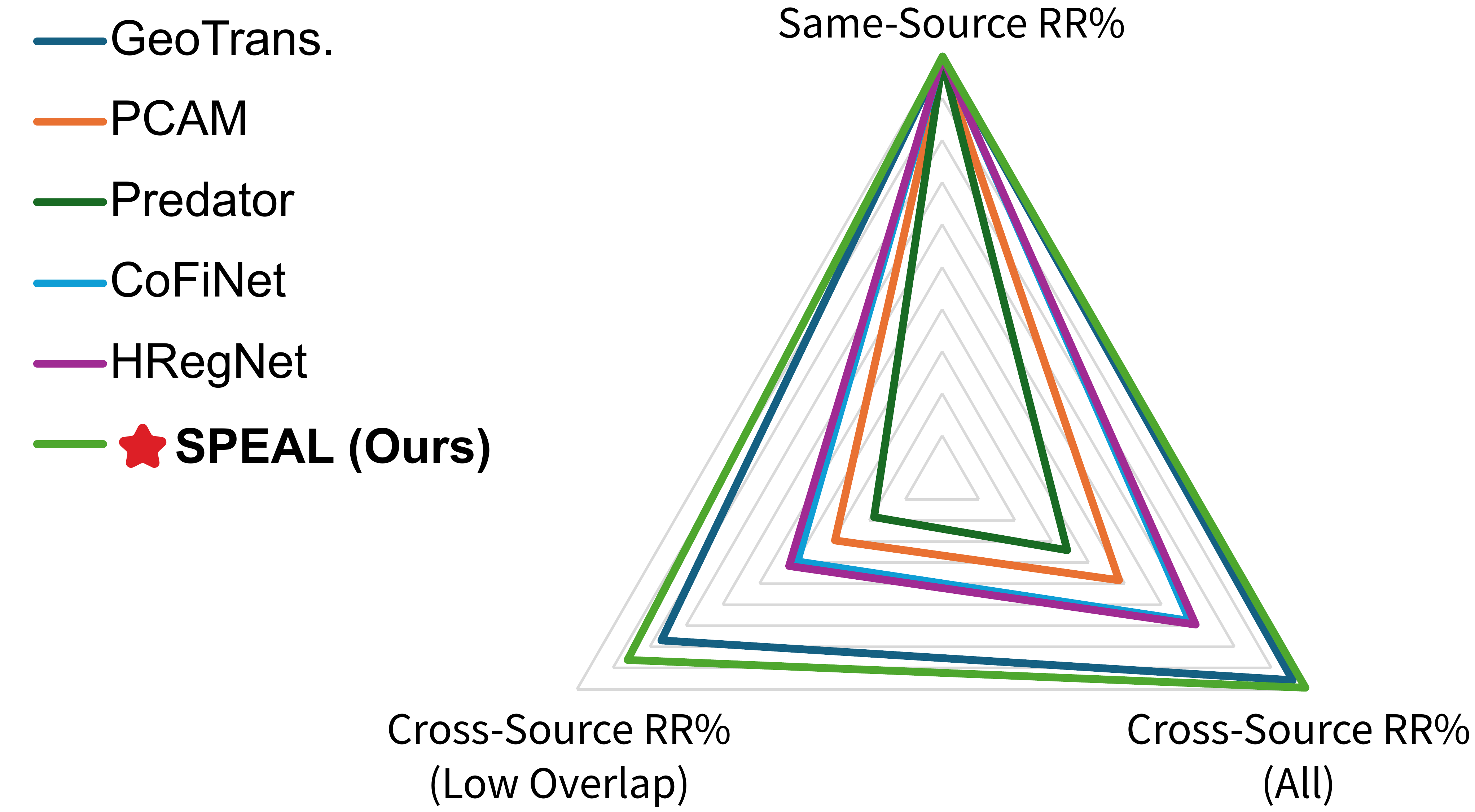}
  \caption{\textbf{Impossible Triangle of Current Methods.} Registration recalls under different settings are shown: KITTI Odometry (Same-Source), \kittics{} and the low-overlap test split of \kittics{}. Existing methods fail to perform as well as \mynetwork{} on all three challenging circumstances.}
  \label{fig:teaser}
\end{figure}

Point cloud registration is an essential task in graphics, vision, and robotics.
\qs{It aims at estimating} a rigid transformation \qs{to} align two partially overlapping frames of point clouds. 
\qs{Recently, there has been a surge of interest in learning-based point cloud registration methods. 
These methods have made significant progress in addressing the sparsity, partial overlap, and complex distribution of point clouds in large outdoor scenes \cite{lu2021hregnet,huang2021predator,yew2022regtr,qin2022geometric}. 
However, the practical application and advances in point cloud acquisition present more challenges for point cloud registration, including unstructured scenes and cross-source data.}

In \emph{unstructured scenes}, the complex natural scenes and objects often make it difficult to learn discriminative features for registration.
This results in degraded performance of registration algorithms.
In the case of \emph{cross-source data}, challenges mainly arise from partial overlap, as well as considerable differences in scale and density, leading to difficulties in effective feature matching. 
The combination of noise and outliers from different sources further downgrades the quality of correspondences.
Existing methods either focus solely on same-source point clouds, or overlook the intrinsic topological natures of the point clouds.
This leads to suboptimal results for challenging scenarios such as cross-source point cloud registration and unstructured scenes.

We have observed that skeletons serve as an efficient and robust geometric representation for point clouds, exhibiting significant potential in various point cloud understanding tasks \cite{shi2021skeleton,lin2021point2skeleton}.
They can effectively encode the geometric intricacy of point clouds.
Inspired by this, we propose a novel transformer-based approach, termed \mynetworklongbf{} (\mynetwork{}), to address the aforementioned challenges. 
Our method utilizes skeletal geometric priors to learn discriminative features for accurate and robust correspondences.
To the best of our knowledge, our approach is the first to facilitate point cloud registration with skeletal geometric priors. 
Such skeletal geometric priors \qs{encourage} robust feature learning by explicitly encoding the intrinsic topological characteristics, thereby facilitating the correspondences and registration results, 
as shown in Fig.~\ref{fig:teaser}

Specifically, to incorporate skeletal representations as a geometric prior, \mynetwork{} comprises three key components: \emph{\semfullbf{} (\sem)}, \emph{\sagtrfullbf{} (\sagtr)}, and \emph{\cdsfullbf{} (\cds)}. 
\qs{First, with the insights from the medial axis transform (MAT) \cite{blum1967transformation}, \sem{} extracts a set of skeleton points with skeletal features from input point clouds} in an unsupervised manner. 
It is robust to noise and density variances. 
Next, \sagtr{} is designed to learn skeleton-aware discriminative features, facilitating accurate and robust correspondences.
It explicitly captures the topological natures and effectively learns inter-point-cloud geometric correlations with \qs{our} skeletal representations. 
Finally, \cds{} samples reliable correspondences from both superpoints and skeleton points, which produces reliable coarse correspondences with awareness of the skeletal structure.

Extensive experiments are carried out on two datasets. One is \emph{KITTI Odometry}, a large-scale outdoor registration benchmark \cite{geiger2012we}, as well as its cross-source variant named \emph{\kittics} proposed by us.
The other is a large-scale cross-source dataset mostly consisting of unstructured forest scenes \cite{weiser2022individual}.
The results demonstrate that \mynetwork{} is effective and robust for both same-source and cross-source point cloud registration, as well as for point clouds of unstructured scenes.

Overall, our contributions are threefold:
\begin{itemize}
	\item We propose a novel learning-based point cloud registration approach, \mynetwork, which is the first to utilize skeletal representations as a geometric prior to achieve improved performance.
	\item The proposed \sem{} is an effective and portable skeleton extractor. Our \sagtr{} combined with \cds{} effectively produces accurate and robust correspondences for both same-source and cross-source point cloud registration. 
	\item \kittics{}, a novel cross-source point cloud dataset, meets the dire need~\cite{huang2021comprehensive} 
    \qs{of cross-source point cloud registration benchmarks.} 
    This opens up the possibility to bridge the gap between sensor technology and cross-source applications.
\end{itemize}

\section{Related Work}



\customparagraph{Learning-based Registration Methods}
Learning-based registration methods fall into two categories: correspondence-based methods and direct registration methods. 
Correspondence-based methods \cite{choy2019fully,deng2018ppf,deng2018ppfnet,gojcic2019perfect,yao2020quasi} first extract correspondences between two point clouds, and then estimate the transformation with robust pose estimators. 
However, traditional robust estimators suffer from slow convergence and are sensitive to outliers. 
To address this, deep robust estimators \cite{choy2020deep,bai2021pointdsc,pais20203dregnet,lee2021deep} 
utilize deep neural networks to reject outliers and compute the transformation. While these methods require a training procedure, they improve accuracy and speed. 
Direct registration methods 
directly estimate the transformation between two point clouds 
in an end-to-end way. Inspired by Iterative Closest Point \cite{besl1992method}, some of them \cite{fu2021robust,wang2019prnet,wang2019deep,yew2020rpm} iteratively build soft correspondences and then estimate the transformation
with SVD.
Others \cite{xu2021omnet,aoki2019pointnetlk,huang2020feature} extract a global feature vector and regress the transformation directly with a neural network. 
However, such methods could potentially fail in large-scale scenes.

\customparagraph{Transformers in Point Cloud Registration}
Originally designed for NLP tasks, Transformers \cite{vaswani2017attention} have shown remarkable efficacy in computer vision \cite{misra2021end,carion2020end,dosovitskiy2020image,yu2021pointr}. 
Recently, transformer-based methods for point cloud registration have also emerged.
Geometric Transformer \cite{qin2022geometric} leverages transformer layers for superpoint matching,
while REGTR \cite{yew2022regtr} uses a transformer cross-encoder and a transformer decoder to directly predict overlap scores.
PEAL\cite{yu2023peal} leverages additional overlap priors from 2D images. 

\customparagraph{Point Cloud Skeletal Representations}
\qs{The curve skeleton is a widely-used skeletal representation due to its simplicity \cite{huang2013l1,ma2003skeleton,au2008skeleton,cao2010point}. 
It has shown its potential in some learning-based methods \cite{xu2019predicting,shi2021skeleton} like keypoint extraction.}
However, it is only well-defined for tubular geometries, thus limiting its
expressiveness for point clouds with complex shapes or in large-scale scenes.
The \textbf{M}edial \textbf{A}xis \textbf{T}ransform (MAT) \cite{blum1967transformation} is another skeletal representation capable of encoding arbitrary shapes.
\qs{Some methods \cite{sun2015medial,yan2018voxel,li2015q} employ simplification techniques to alleviate the distortion caused by surface noise, but they are computationally ineffective and require watertight input surfaces.}
Recent learning-based efforts \cite{lin2021point2skeleton,wen2023learnable} use deep neural networks to predict MAT-based skeletons, thus greatly enhancing the robustness and computational efficiency.
These methods have shown promising results in various 3D vision tasks, including shape reconstruction and point cloud sampling \cite{wen2023learnable}.

\section{Method}

\begin{figure*}[ht]
  \centering
  \includegraphics[width=0.9\linewidth]{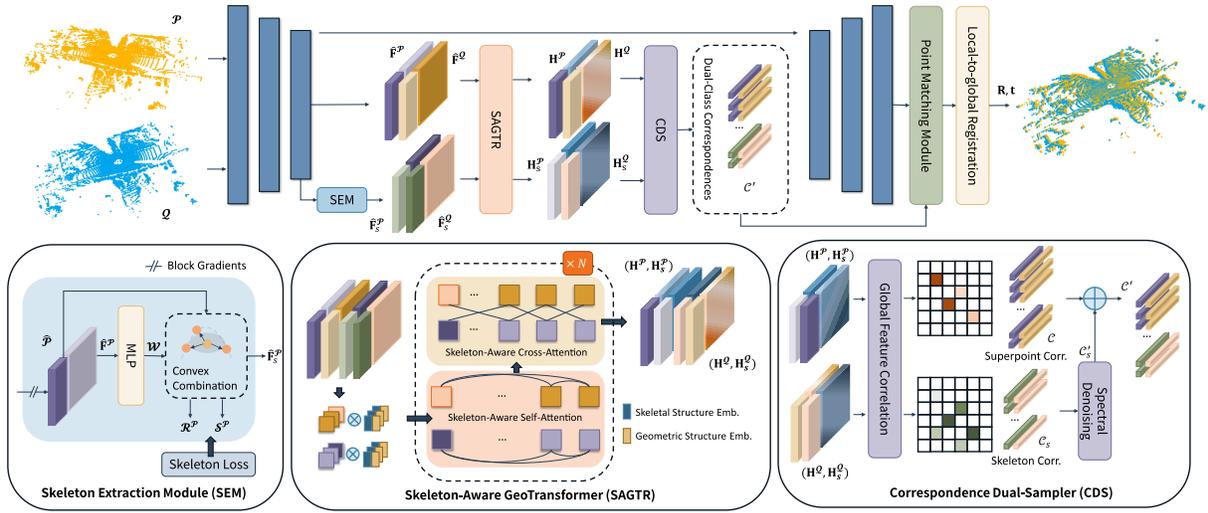}
  \caption{\textbf{The Overall Pipeline of \mynetwork.} The backbone extracts superpoints and multi-level features from $\mathcal{P}$ and $\mathcal{Q}$. Then, SEM and SAGTR extract skeletal representations and learn discriminative skeleton-aware features, respectively. Finally, CDS extracts hybrid coarse correspondences with skeletal priors. The result transformation is computed with LGR. }
  \label{fig:pipeline}
\end{figure*}

\customparagraph{Problem Statement}
Given two point clouds $\mathcal{P}=\{\mathbf{p}_i \in \mathbb{R}^3 | i=1,\ldots,N\}$ and 
$\mathcal{Q}=\{\mathbf{q}_i \in \mathbb{R}^3 | i=1,\ldots,M\}$, \qs{our goal is} 
to align the two point clouds by estimating a rigid transformation $\mathbf{T}=\{\mathbf{R},\mathbf{t}\}$, 
where $\mathbf{R} \in SO(3)$ is a 3D rotation matrix and $\mathbf{t} \in \mathbb{R}^3$ is a 3D translation vector.
\qs{The transformation can be solved by:}
\begin{equation}
\label{eq:transformation}
\min_{\mathbf{R},\mathbf{t}}\sum\nolimits_{(\mathbf{p}_{x_i},\mathbf{q}_{y_i})\in\mathcal{C}^\star} \|\mathbf{R}\mathbf{p}_{x_i}+\mathbf{t}-\mathbf{q}_{y_i}\|^2,
\end{equation}
where $\mathcal{C}^\star$ denotes the set of correspondences between two point clouds $\mathcal{P}$ and $\mathcal{Q}$.
In reality, $\mathcal{C}^\star$ is usually unknown.
Hence, we need to establish accurate correspondences $\mathcal{C}$ between two point clouds for a good transformation.


\customparagraph{Overview \xkznew{and Notations}}
\qs{Our work} leverages skeletal priors in an end-to-end neural network to facilitate correspondences.
The pipeline is shown in Fig.~\ref{fig:pipeline}, following the hierarchical correspondence paradigm.
To extract multi-level features for point clouds, we leverage the KPConv-FPN backbone \cite{lin2017feature,thomas2019kpconv}.
\xkznew{The points at the coarsest level of the backbone are \emph{superpoints}, denoted as $\hat{\mathcal{P}}$ and $\hat{\mathcal{Q}}$. Their associated features  are $\hat{\mathbf{F}}^\mathcal{P}\in\mathbb{R}^{|\hat{\mathcal{P}}|\times d_t}$ and $\hat{\mathbf{F}}^\mathcal{Q}\in\mathbb{R}^{|\hat{\mathcal{Q}}|\times d_t}$.}
Then, our proposed \sem, \sagtr{} and \cds{} are used to extract reliable and accurate coarse correspondences with skeletal priors.
Finally, we employ \xkznew{the Point Matching Module and Local-to-Global Registration \cite{qin2022geometric} to obtain dense correspondences and} estimate the final rigid transformation.

\subsection{\semfullcap}

\xkznew{The \semfullcap{} aims to approximate the Medial Axis Transform (MAT) by leveraging a convex combination of input points, which provides a well-defined skeletal representation for arbitrary shapes } in an unsupervised manner.
Inspired by existing methods \cite{lin2021point2skeleton}, it overcomes the computational expense and sensitivity to surface noise of traditional MAT computation. 

Specifically, for all points in \xkznew{$\hat{\mathcal{P}}\in\mathbb{R}^{|\hat{\mathcal{P}}|\times 3}$ and their features $\hat{\mathbf{F}}^\mathcal{P}\in\mathbb{R}^{|\hat{\mathcal{P}}|\times d_t}$}, 
\sem{}  aims to extract \qs{$N_s$} skeleton points $\mathcal{S}^\mathcal{P}\in\mathbb{R}^{N_s\times3}$, their skeletal features $\hat{\mathbf{F}}^\mathcal{P}_s\in\mathbb{R}^{N_s\times d_t}$, and their radii $\mathcal{R}^\mathcal{P}\in\mathbb{R}^{N_s\times 1}$ .
We extract skeletons for $\mathcal{Q}$ in the same way.
\qs{To this end,}
\xkznew{we employ a multi-layer perceptron (MLP) to predict the weights $\mathbf{\mathcal{W}}\in\mathbb{R}^{|\hat{\mathcal{P}}|\times N_s}$.
The MLP is shared across $\hat{\mathcal{P}}$ and $\hat{\mathcal{Q}}$.}
Then, the skeleton points $\mathcal{S}^\mathcal{P}$ are obtained as the convex combination\cite{lin2021point2skeleton} of input points $\hat{\mathcal{P}}$:
\begin{equation}
\label{eq:convexs}
\mathcal{S}^\mathcal{P}=\mathbf{\mathcal{W}}^T\hat{\mathcal{P}} \ s.t.\  j\!=\!1,\ldots,N_{\qs{s}}, \ \sum\nolimits_{i=1}^{|\hat{\mathcal{P}}|}\mathcal{W}(i,j)=1.
\end{equation}
The weighting scheme enhances the robustness of skeleton extraction by effectively filtering out noise and outliers. Similarly, we extract their skeletal features by $\hat{\mathbf{F}}^\mathcal{P}_s = \mathbf{\mathcal{W}}^T\hat{\mathbf{F}}^{\qs{\mathcal{P}}}$.

\qs{To predict the radius of each skeleton point, we first compute the closest distance for an input point $\hat{\mathbf{p}}$ to all skeleton points as follows:}
\begin{equation}
  d(\hat{\mathbf{p}},\mathcal{S}^\mathcal{P})=\min\nolimits_{\mathbf{s}\in\mathcal{S}^\mathcal{P}}\|\hat{\mathbf{p}}-\mathbf{s}\|_2.
\end{equation}
The distances for all input points are then summarized in a vector $\mathcal{D}^\mathcal{P}\in\mathbb{R}^{|\hat{\mathcal{P}}|\times 1}$.
Next, the radii of all the skeleton points are computed through a linear combination of their closest distances from all the input points, \ie $\mathcal{R}^\mathcal{P}=\mathcal{W}^T\mathcal{D}^\mathcal{P}$.
This approximation is based on the observation that the predicted weights for a skeleton point $\mathbf{s}$ are significant only for the input points that in a local neighborhood of $\mathbf{s}$,
and diminish to 0 for the input points far away from $\mathbf{s}$. 

\qs{The skeleton extraction is a fundamentally different task from the point cloud registration.} 
Therefore, the module is separately supervised by the \emph{skeleton loss} \qs{\cite{lin2021point2skeleton}}, and we block the gradient flow from this module to the backbone \xkznew{for a more stable training process and better performance (See supplementary materials).}

\subsection{\sagtrfullcap}

The registration of cross-source point clouds poses significant challenges, including noise, density differences, and scale variances. 
\qs{Skeleton points exhibit consistency and robustness against these challenges. 
Therefore, we propose the \sagtr{} module to encode the structure of point clouds. 
It comprises two key components: \emph{Skeleton-aware Geometric Self-Attention} and \emph{Skeleton-aware Cross-Attention}. }
They are interleaved for $N_t$ times to further extract non-skeletal and skeletal hybrid features $(\mathbf{H}^\mathcal{P}, \mathbf{H}_s^\mathcal{P})$ and $(\mathbf{H}^\mathcal{Q}, \mathbf{H}_s^\mathcal{Q})$.
\xkznew{These features encode inter-point-cloud and intra-point-cloud correlations and skeletal geometric priors.}
They contribute to accurate and robust coarse correspondences.

\begin{figure}[t]
  \centering
  \includegraphics[width=\linewidth]{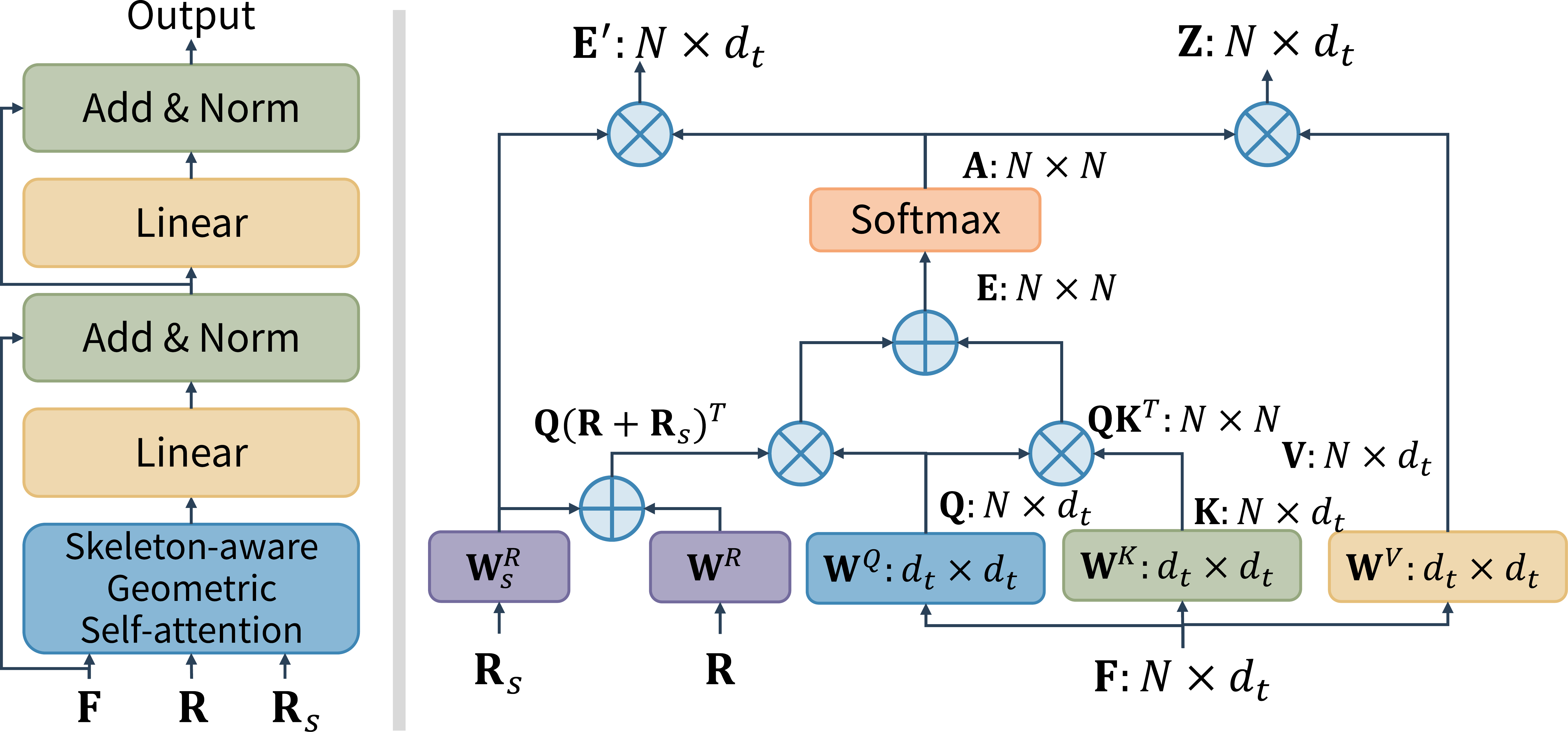}
  \caption{
  \qs{The structure (left) and computational graph (right) of skeleton-aware geometric self-attention.}
  }
  \label{fig:sgsa}
\end{figure}

\customparagraph{Skeleton-Aware Geometric Self-Attention} 
In the following, we describe the computation for $\hat{\mathcal{P}}$, and the computation for $\hat{\mathcal{Q}}$ is exactly the same.
\xkznew{Given an input feature matrix $\mathbf{X}\in\mathbb{R}^{L\times d_t}$ ($L=|\hat{\mathcal{P}}|+N_s$ is the length of the input sequence), }
the output feature matrix $\mathbf{Z}\in\mathbb{R}^{L\times d_t}$ is the 
weighted sum of all projected input features:

\begin{equation}
\label{eq:z}
\mathbf{z}_i=\sum\nolimits_{j=1}^{\qs{L}} a_{i,j} (\mathbf{x}_j\mathbf{W}^V),
\end{equation} 
where $a_{i,j}$ is the weight coefficient computed by a row-wise softmax on 
the attention score $e_{i,j}$, and $e_{i,j}$ is computed as:
\begin{equation}
\label{eq:e}
e_{i,j}=(\mathbf{x_i}\mathbf{W}^Q)(\mathbf{x_j}\mathbf{W}^K+\mathbf{r}^p_{i,j}\mathbf{W}^P+\mathbf{r}^s_{i,j}\mathbf{W}^S)^T / \sqrt{d_t}.
\end{equation}
Here, \qs{$\mathbf{r}^s_{i,j}$ and $\mathbf{r}^p_{i,j}$} are \emph{Skeleton-Aware Structure Embedding} and \emph{Point-Wise Structure Embedding} 
respectively. 
\qs{We follow \cite{qin2022geometric} to compute $\mathbf{r}^p_{i,j}$, which encodes non-skeletal geometric structures between superpoints. $\mathbf{r}^s_{i,j}$ encodes skeletal latent geometric information of point clouds, which will be described next.}
\xkznew{$\mathbf{W}^Q ,\mathbf{W}^K,\mathbf{W}^V ,\mathbf{W}^P, \mathbf{W}^S \in \mathbb{R}^{d_t\times d_t}$} are the respective projections for queries, keys, values, 
point-wise structure embedding and skeleton-aware structure embedding. 
Fig.~\ref{fig:sgsa} shows the computation of Skeleton-aware Geometric Self-attention.

We design a novel approach, termed \emph{Skeleton-Aware Structure Embedding}, to encode \qs{skeletal} latent structural information in the geometric space.
\xkznew{The insight is to leverage the}
transformation invariance and robustness 
in the local geometric structure formed by the skeleton points.
\qs{This embedding includes \emph{skeleton-wise distance embedding} and \emph{skeleton-wise angular embedding}.}
They respectively capture distance and angle information of the local geometric structure formed by skeleton points around superpoints.

\begin{figure}[t]
  \centering
  \includegraphics[width=\linewidth]{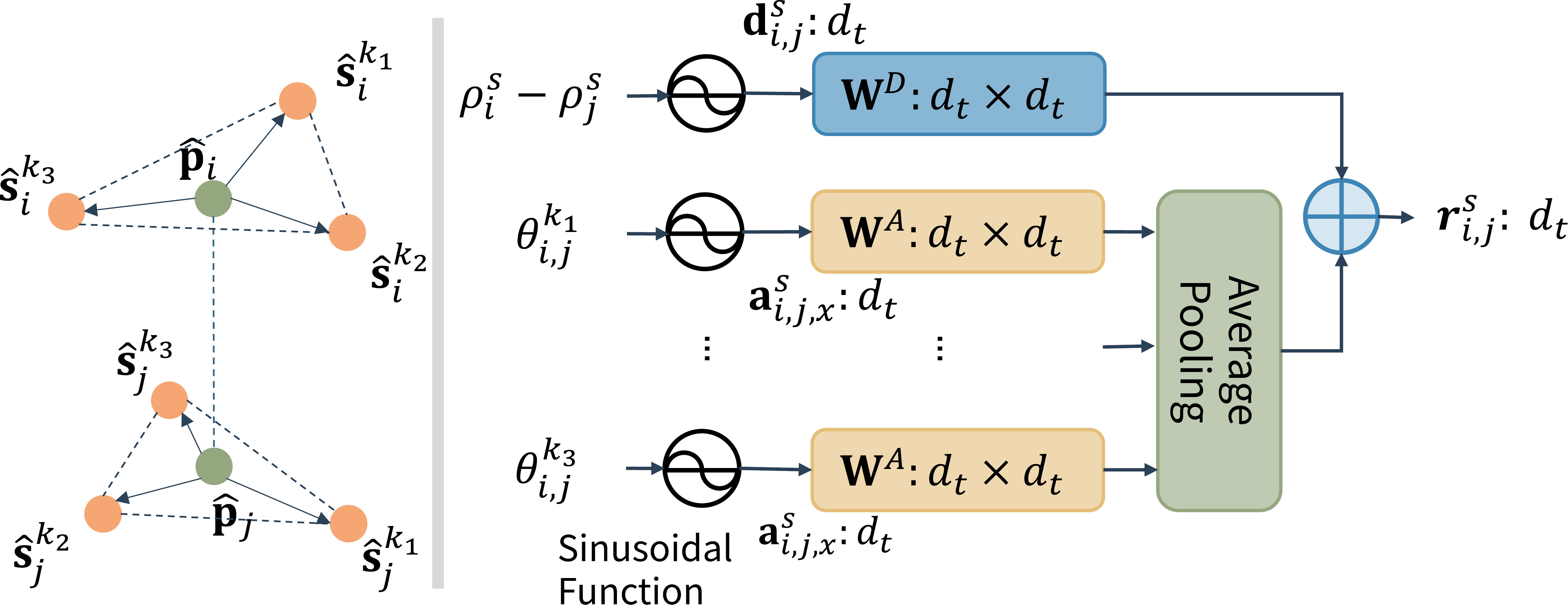}
  \caption{The computation of skeleton-aware structure embedding. }
  \label{fig:sse}
\end{figure}

Specifically, given two superpoints $\hat{\mathbf{p}}_i,\hat{\mathbf{p}_j}\in\hat{\mathcal{P}}$, their \textit{k}NN skeleton points are $\mathbf{s}^i_1,\ldots,\mathbf{s}^i_k\in\mathcal{K}^s_i$ and $\mathbf{s}^j_1,\ldots,\mathbf{s}^j_k\in\mathcal{K}^s_j$, respectively. 
Based on them, as shown in Fig.~\ref{fig:sse}, the computation of skeleton-wise structure embedding $\mathbf{r}^s_{i,j}$ is twofold: 
\qs{1) \emph{ Skeleton-Aware Distance Embedding.} For each superpoint $\mathbf{p}_j$, we first compute $\rho^s_j=\sum_{\mathbf{s}^j_x\in\mathcal{K}_j^s}d(\mathbf{s}^j_x,\mathbf{p}_j)$ , where $d(\mathbf{s}^j_x,\mathbf{p}_j)=\|\mathbf{s}^j_x-\mathbf{p}_j\|_2$ denotes the distance between $s_i$ and $p_j$ in the Euclidean space. Then, the skeleton-aware distance embedding $\mathbf{d}^{s}_{j}$ is computed by applying a sinusoidal function on $(\rho^s_i-\rho^s_j)/\sigma^s_d$. 
2) \emph{Skeleton-Aware Angular Embedding.} For each skeleton point $s^j_x\in\mathcal{K}^s_j$, we first compute the angle $\theta^x_{i,j}=\angle(\mathbf{s}^j_x-\mathbf{p}_j,\mathbf{p}_i-\mathbf{p}_j)$.
Based on the angles, the skeleton-wise angular embedding $\mathbf{a}^{s}_{i,j,x}$ is computed by applying a sinusoidal function on $\theta^x_{i,j}/\sigma^s_a$.}
\qs{Herein,} $\sigma^s_d$ and $\sigma^s_a$ control the sensitivity on skeleton-wise distances and angles respectively. 
The final skeleton-aware structure embedding $\mathbf{r}^s_{i,j}$ is \xkznew{the aggregation of the skeleton-\qs{aware} angular embedding $\mathbf{a}^{s}$ and the skeleton-\qs{aware} distance embedding $\mathbf{d}^{s}$:}
\begin{equation}
\label{eq:rs}
\mathbf{r}^s_{i,j}=\mathbf{d}^{s}_{i,j}\mathbf{W}^D+\mathrm{mean}_x\{\mathbf{a}^{s}_{i,j,x}\mathbf{W}^A\}, 
\end{equation} 
\xkznew{where $\mathbf{W}^D$ and $\mathbf{W}^A$ are trainable weights. }

To \qs{help} the successive cross-attention layers \qs{to capture the geometric structure} with skeletal priors, 
our skeleton-aware self-attention layers also produce the skeleton-aware positional encoding $\mathbf{E}^{'}$ by applying the attention scores on the skeleton-aware structure embedding $\mathbf{r}^s_{i,j}$:
\begin{equation}
\label{eq:spe}
{\mathbf{E}}^{'}_{i,k} = \sum\nolimits_{j=1}^{\qs{L}} a_{i,j} \cdot r_{i,j,k}^s.
\end{equation}

\begin{figure}[t]
  \centering
  \includegraphics[width=\linewidth]{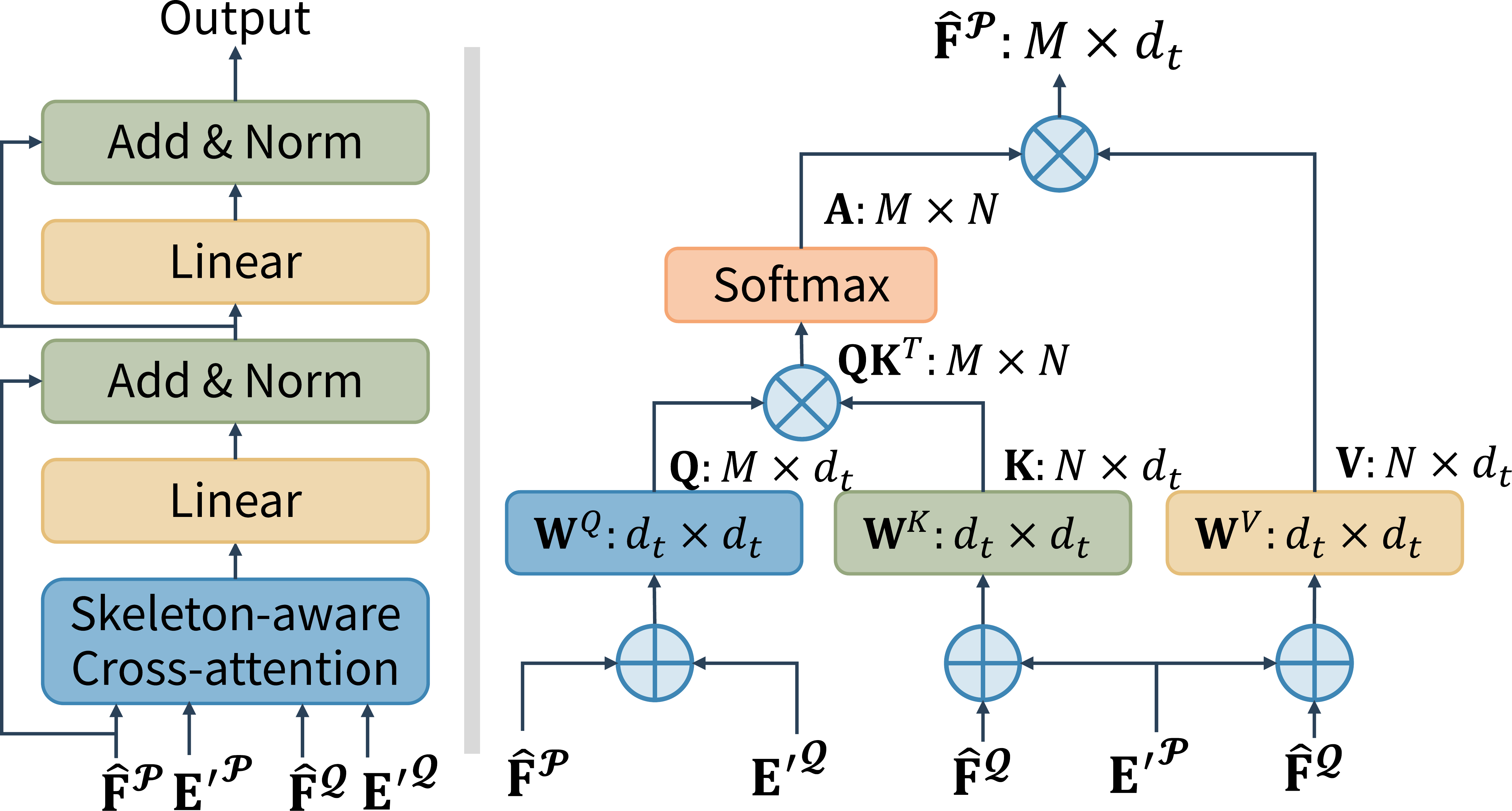}
  \caption{
  \qs{The structure (left) and computational graph (right) of skeleton-aware cross-attention.}
  }
  \label{fig:sgca}
\end{figure}

\customparagraph{Skeleton-Aware Cross-Attention}
Several existing works 
\cite{qin2022geometric,yew2022regtr} have utilized the cross-attention mechanism for inter-point-cloud feature exchange.
However, they either lack positional encoding or fail to explicitly consider the geometric structure of point clouds, 
leading to suboptimal performance.
To address this, we propose the skeleton-aware cross-attention to explicitly learn the correlation of point clouds with skeletal priors, as is shown in Fig.~\ref{fig:sgca}.

Given feature maps with their skeleton-aware positional encoding $(\mathbf{X}^\mathcal{P},\mathbf{E}^{'}_\mathcal{P})$ and $(\mathbf{X}^\mathcal{Q},\mathbf{E}^{'}_\mathcal{Q})$ for $\hat{\mathcal{P}}$ and  $\hat{\mathcal{Q}}$ respectively. 
A skeleton-aware cross-attention layer first adds the positional encoding to features to produce skeleton-aware features $\mathbf{X}^{'\mathcal{P}}$ and \qs{$\mathbf{X}^{'\mathcal{Q}}$}.
Then, the output for $\hat{\mathcal{P}}$ are computed with the features of $\hat{\mathcal{Q}}$:
\begin{equation}
\label{eq:cross-attention}
\mathbf{z}_i^\mathcal{P} = \sum\nolimits_{j=1}^{|\hat{\mathcal{Q}}|} a_{i,j}(\mathbf{x}_j^{'\mathcal{Q}}\mathbf{W}^V).
\end{equation}
Similarly, the weights $a_{i,j}$ are computed by a row-wise softmax on the attention score $e_{i,j}$:
\begin{equation}
\label{eq:cross-attention-score}
e_{i,j} = (\mathbf{x}_i^{'\mathcal{P}}\mathbf{W}^Q)(\mathbf{x}_j^{'\mathcal{Q}}\mathbf{W}^K)^T / \sqrt{d_t}.
\end{equation}
\qs{The same cross-attention implementation goes for $\hat{\mathcal{Q}}$.} 
In contrast to skeleton-aware geometric self-attention \qs{that} captures the intra-point-cloud transformation-invariant geometric structure, 
the cross-attention \qs{here} captures the inter-point-cloud geometric corrections and consistency.
The hybrid features obtained from \sagtr{} are therefore discriminative enough for matching.

\subsection{\cdsfullcap}

\xkznew{With} discriminative features, it is vital to extract accurate coarse correspondences. 
Geometric Transformer \cite{qin2022geometric} only matches the superpoints.
\xkznew{Despite its efficiency}, superpoints are sparse and may be unrepeatable, \xkznew{leading to outlier correspondences.}
Existing efforts strive to tackle \xkznew{this issue with} sophisticated
sampling strategies \cite{li2023hybridpoint} or overlap priors from 2D images \cite{yu2023peal}.
\qs{Unfortunately, their required complicated sampling and extra 2D images result in suboptimal computational efficiency, which hinders the application of such methods.}
\xkznew{To this end, we propose \cds{} to effectively augment the correspondence set with our effective skeletal representation, leading to a more accurate and robust hybrid coarse correspondence set.}


\xkznew{We separately construct the non-skeletal correspondence set $\mathcal{C}$ and skeletal correspondence set  $\mathcal{C}_s$ by feature matching:} 
We first compute the Gaussian correlation matrix $\mathbf{S}\in\mathbb{R}^{|\hat{\mathcal{P}}|\times|\hat{\mathcal{Q}}|}$ \xkznew{for the normalized features $\mathbf{F}^\mathcal{P}$ and $\mathbf{F}^\mathcal{Q}$}, and then
use a dual-normalization operation \cite{sun2021loftr,rocco2018neighbourhood} to suppress ambiguous matches.
Finally, we select at most $N_c$ largest entries for each correspondence set.

\qs{Since} skeleton points \xkznew{lying} on non-overlap regions \xkznew{may introduce outlier correspondences}, we introduce the \emph{Spectral \qs{Denoising}} procedure to filter $\mathcal{C}_s$ \xkznew{with a spectral matching algorithm \cite{leordeanu2005spectral}:}
We \xkznew{firstly} compute a compatibility matrix based on the 3D spatial consistency of $\mathcal{C}_s$.
\xkznew{Then,} we iteratively remove components conflicting with the item of the maximum principal eigenvector until either the principal eigenvector becomes zero or $|\mathcal{C}_s|$ equals the minimum number of the main cluster.
The main cluster, denoted as $\mathcal{C}_s^{'}$, is the final skeletal correspondence set. 

Finally, \xkznew{we resample} the least confident $N_s$ entries of $\mathcal{C}$ with top $N_k$ correspondences in $\mathcal{C}_s^{'}$ to obtain the hybrid correspondence set $\mathcal{C}^{'}$, \xkznew{thereby improving} the accuracy and robustness of the hybrid coarse correspondence set by \xkznew{replacing} potential outliers with more reliable correspondences.

\subsection{Losses}

We use a registration loss and a skeleton loss to supervise \mynetwork.
The registration loss consists of \emph{Overlap-aware Circle Loss} ($\mathcal{L}_{oc}$) and \emph{Point Matching Loss} ($\mathcal{L}_p$) from Geometric Transformer \cite{qin2022geometric}:
\begin{equation}
\begin{aligned}
\mathcal{L} &= \mathcal{L}_{oc} + \mathcal{L}_{p}.
\end{aligned}
\end{equation}
\qs{The skeleton loss in \citet{lin2021point2skeleton} is used to supervise the \sem.}
\qs{It} is the weighted sum of \emph{Sampling Loss} $\mathcal{L}_{\text{s}}$, \emph{Point-to-sphere Loss} $\mathcal{L}_{\text{r}}$ and \emph{Radius Regularizing Loss} $\mathcal{L}_{\text{p2s}}$:
\begin{equation}
\label{eq:chamfer}
\begin{aligned}
\mathcal{L}_{\text{skeleton}}&=\mathcal{L}_{\text{s}}+\lambda_1\mathcal{L}_{\text{p2s}}+\lambda_2\mathcal{L}_{\text{r}},
\end{aligned}
\end{equation}
\xkznew{where $\lambda_1$ and $\lambda_2$ are hyperparameters to balance the losses. Please refer to the supplementary material for more details.}

\section{Experiments}

\subsection{Datasets and Experimental Setup}


\customparagraph{Same-Source Dataset}
\textcolor{purple}{The KITTI Odometry dataset \cite{geiger2012we} serves as a widely-used dataset for odometry and SLAM evaluation. It can also be employed to test same-source point cloud registration. 
This dataset comprises 11 sequences of LiDAR point clouds.}
We follow the existing practices \cite{qin2022geometric,huang2021predator} to use sequences 00-06 for training, sequences 07-08 for evaluation and sequences 09-10 for testing.

\customparagraph{Cross-Source Datasets} 
Currently, there are few cross-source datasets \qs{of} large-scale outdoor scenes available for registration tasks.
This hinders the development of cross-source registration methods.
Therefore, we have developed a novel dataset\footnote{The dataset will be made publicly available} termed \emph{\kittics{}} derived from KITTI Odometry. 
Our proposed dataset includes 11 sequences of LiDAR point clouds and reconstructed point clouds generated from stereo images using MonoRec \cite{wimbauer2021monorec}. \xkznew{We improve the reconstruction quality with a filter-and-combine strategy. Please refer to the supplementary material for details.}

\textcolor{purple}{The \germanyforest{} dataset is \xkznew{derived from} an existing large-scale forest scene dataset \cite{weiser2022individual}. 
It contains cross-source point cloud data acquired in 12 forest plots in south-west Germany under leaf-on and leaf-off conditions.
Each plot provides Airborne Laser Scanning (ALS), Terrestrial Laser Scanning (TLS) and UAV-borne Laser Scanning (ULS) point clouds. In this paper, we use ALS and ULS scans to evaluate the cross-source registration performance. 
In experiments, we use 10 plots for training, 1 for validation and 1 for testing. }


\customparagraph{Data Preprocessing} 
For the \germanyforest{} dataset, the point clouds of each plot are subdivided into $30\mathrm{m}\times30\mathrm{m}\times30\mathrm{m}$ blocks to make \qs{them} suitable for the registration task.
For all datasets, the Iterative Closest Point (ICP) algorithm from the Open3D library \cite{zhou2018open3d} is used to refine the noisy ground truth transformation, following previous works \cite{qin2022geometric,lu2021hregnet}.
The point clouds are  downsampled with a voxel size of $0.3\mathrm{m}$.

\customparagraph{Metrics} 
\qs{Following previous practices \cite{qin2022geometric,lu2021hregnet,huang2021predator}, we evaluate the registration performance using following metrics:}
\emph{Relative Rotation Error} (RRE), \emph{Relative Translation Error} (RRE), and 
\emph{Registration Recall} (RR). 
We use a RRE threshold and a RTE threshold to compute RR for all datasets ($\textrm{RRE}<0.5\degree$ and $\textrm{RTE}<0.3\textrm{m}$ for \germanyforest{} and $\textrm{RRE}<5\degree$ and $\textrm{RTE}<2\textrm{m}$ for KITTI datasets). 
Additionally, we measure the quality of correspondences with Inlier Ratio (IR), which is the fraction of extracted correspondences whose residuals are below a certain threshold under the ground-truth transformation.

\customparagraph{Implementation Details}
To train \mynetwork{}, we use Adam \cite{kingma2014adam} optimizer with an initial learning rate of $1e-4$ and a weight decay of $1e-6$. 
We train \mynetwork{} for $200$ epochs with a batch size of $1$ on a NVIDIA RTX 3090 GPU.

\customparagraph{Baselines} 
We compare our method with state-of-the-art methods of three classes: 
(a) Traditional methods, including RANSAC \cite{fischler1981random} and FGR \cite{zhou2016fast}.
(b) Transformer-based methods, including CoFiNet \cite{yu2021cofinet}, Predator\cite{huang2021predator}, PCAM \cite{cao2021pcam}, REGTR \cite{yew2022regtr} and Geometric Transformer (abbreviated as GeoTrans.) \cite{qin2022geometric}.
(c) Other learning-based methods, including FCGF \cite{choy2019fully}, DGR \cite{choy2020deep} and HRegNet \cite{lu2021hregnet}.


\subsection{Cross-Source Results}

\begin{table}[t]
	\centering
    \small
    \setlength{\tabcolsep}{2.5pt}
	\begin{tabular}{l|c|c|c|c|c|c}
		\hline
    \multirow{2}{*}{Method} & \multicolumn{3}{c|}{ \kittics } & \multicolumn{3}{c}{KITTI Odometry}  \\
    \cline{2-7}
    & RRE($\degree$) & RTE($\textrm{m}$) & RR($\%$) &  RRE($\degree$)  & RTE($\textrm{m}$) & RR($\%$)\\
    \hline
    RANSAC   & 6.14  & 9.46  & 0.8  & 0.54  & 0.13  & 91.9 \\
    FGR  & -- & -- & --  & 0.96  & 0.93  & 39.4\\
    \hline
    FCGF  & --  &  --  &  --  &  0.30 & 0.095 & 96.6 \\
    DGR &  --  &  -- &  --  & 0.37 & 0.320 & 98.7\\
    HRegNet & 2.19 & 0.84 & 69.3 &  0.29 & 0.120 & 99.7 \\
    \hline
    CoFiNet  & 1.99 & 0.81  & 67.6  & 0.41 & 0.085 & 99.8 \\
    Predator & 5.06 & 2.59  & 34.2 & 0.27 & \textbf{0.068} & 98.8 \\
    PCAM & 4.07 & 2.40 & 45.9 & 0.79 & 0.12 & 98.0 \\
    GeoTrans. & 1.87 & 0.63 & 96.8 & 0.24 & \textbf{0.068} & \textbf{99.8} \\
    \hline
    \mynetwork  & \textbf{1.41} & \textbf{0.58} & \textbf{97.3} & \textbf{0.23} & 0.069 & \textbf{99.8} \\
	  \hline
	\end{tabular}	
  \caption{Cross-source (the proposed \kittics{}) and same-source (KITTI Odometry) registration results on the KITTI datasets. "--" indicates the method is not applicable to the dataset.}
  \label{tbl:kitti}
\end{table}

\customparagraph{\kittics}
The quantitative results are reported in Table~\ref{tbl:kitti}.
Our method achieves state-of-the-art performance on this dataset.
For traditional methods, FGR is not applicable to this dataset and our approach outperforms RANSAC by a large margin. 
HRegNet is a recent SOTA for outdoor large-scale scenes.
However, it presents suboptimal performance on this dataset, 
showing considerable performance decay for cross-source data.
In contrast, our \mynetwork{} is more accurate in terms of all metrics, and has a $28\%$ higher RR than \qs{HRegNet}.
Among transformer-based methods, our method surpasses GeoTrans. by a large margin, showing the effectiveness of the integrated skeletal priors.

\begin{table}[t]
	\centering
    \small
    \setlength{\tabcolsep}{2.5pt}
	\begin{tabular}{l|c|c|c|c|c|c}
		\hline
    \multirow{2}{*}{Overlap} & \multicolumn{2}{c|}{RRE ($\degree$) \textdownarrow} &  \multicolumn{2}{c|}{RTE ($\mathrm{m}$) \textdownarrow}  &  \multicolumn{2}{c}{RR($\%$) \textuparrow}  \\
    \cline{2-7}
    & $\leq 30\%$ &  $> 30\%$ & $\leq 30\%$ &  $> 30\%$ & $\leq 30\%$ &  $>30\%$  \\
    \hline
    RANSAC &  112.0 & 91.1 & 24.66 & 17.4 & -- & -- \\
    FGR &  41.86 &  28.3 & 14.32 & 8.49 &  -- & -- \\
    \hline
    FCGF   & 1.54  & 0.53 & 0.49  & 0.19 &  8.7 & 56.6 \\
    DGR &  1.06 & 0.38 & 0.36 & 0.10 & 32.8 & 76.2 \\
    HRegNet  &  1.16 & 1.40 & 0.141 & 0.238 & 24.7 & 41.7 \\
    \hline
    REGTR &  3.11 & 2.48 & 0.89 & 0.73 & 11.3 & 23.9 \\
    GeoTrans.  & 0.328 & 0.176 & 0.097 & 0.053 & 88.1 & 96.5 \\
    \hline
    \mynetwork\textsubscript{(ours)}& \textbf{0.296} & \textbf{0.165} & \textbf{0.088} & \textbf{0.048} & \textbf{91.7} & \textbf{99.3}\\
	  \hline
	\end{tabular}
  \caption{Cross-source registration results on the \germanyforest{} dataset. "--" indicates that the method is not applicable to the dataset.}
  \label{tbl:gfreg}
\end{table}

\customparagraph{\germanyforest}
\qs{This cross-source dataset is with large scale and unstructured scenes,} 
\qs{which are challenging for registration.}
The \qs{evaluation} results under different overlap ratios are shown in Table~\ref{tbl:gfreg}.  
Traditional methods show suboptimal performance, and RANSAC even fails to register low overlap point clouds.
Learning-based methods overall perform better. However, current methods still suffer from considerable performance decay especially under low overlap condition.
\mynetwork{} outperforms all the others by a large margin, 
showing outstanding robustness introduced by skeletal priors in low overlap and unstructured condition.

\subsection{Same-Source Results}

Table~\ref{tbl:kitti} also \qs{lists} the quantitative results on  the same-source dataset KITTI Odometry.
Compared with recent state-of-the-arts, our method achieves comparable performance in terms of RTE and RR, and outperforms all the other methods in terms of RRE.
This result indicates that our method is also effective for same-source registration, while achieving state-of-the-art performance for cross-source registration.

\subsection{Analysis}

\begin{figure}[ht]
  \centering
  \subfigure[Gorund Truth]{%
  \label{fig:dcgt}%
  \includegraphics[width=0.45\linewidth]{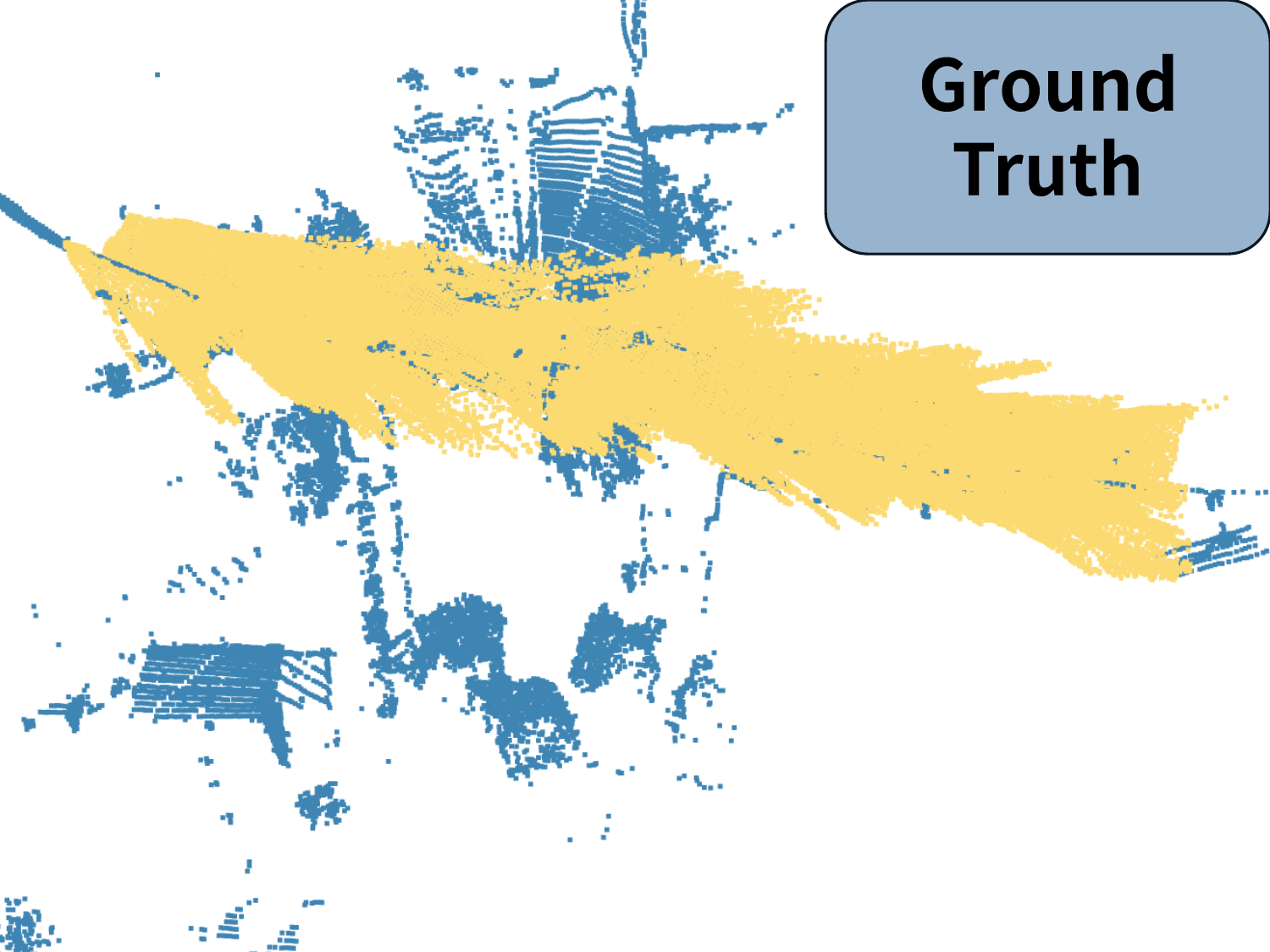}
  }%
  \quad
  \subfigure[Pose (\mynetwork)]{%
  \label{fig:dcpsspeal}%
  \includegraphics[width=0.45\linewidth]{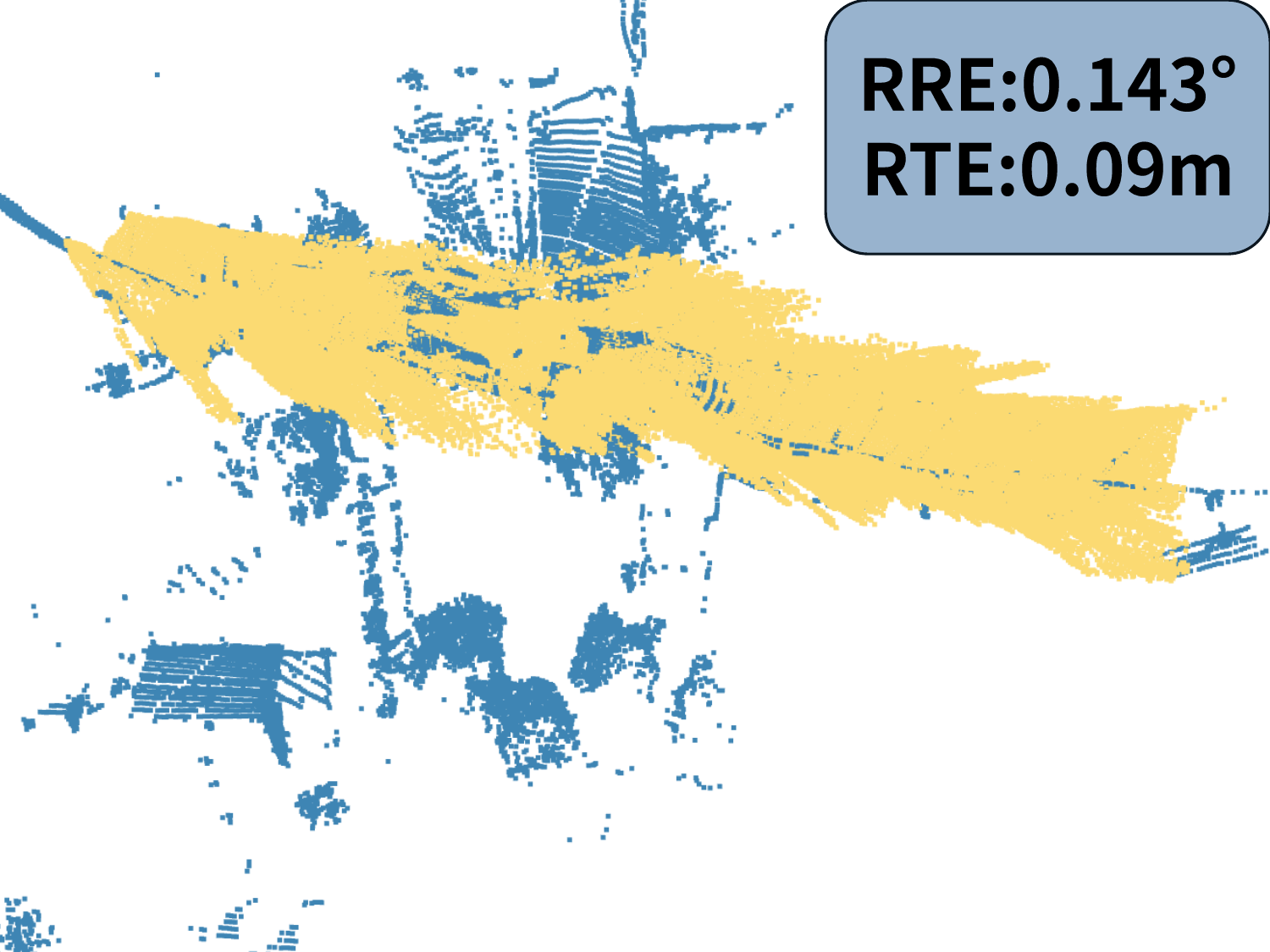}
  }%
  \quad
  \subfigure[Pose (GeoTrans.)]{%
  \label{fig:dcpsgeotrans}%
  \includegraphics[width=0.45\linewidth]{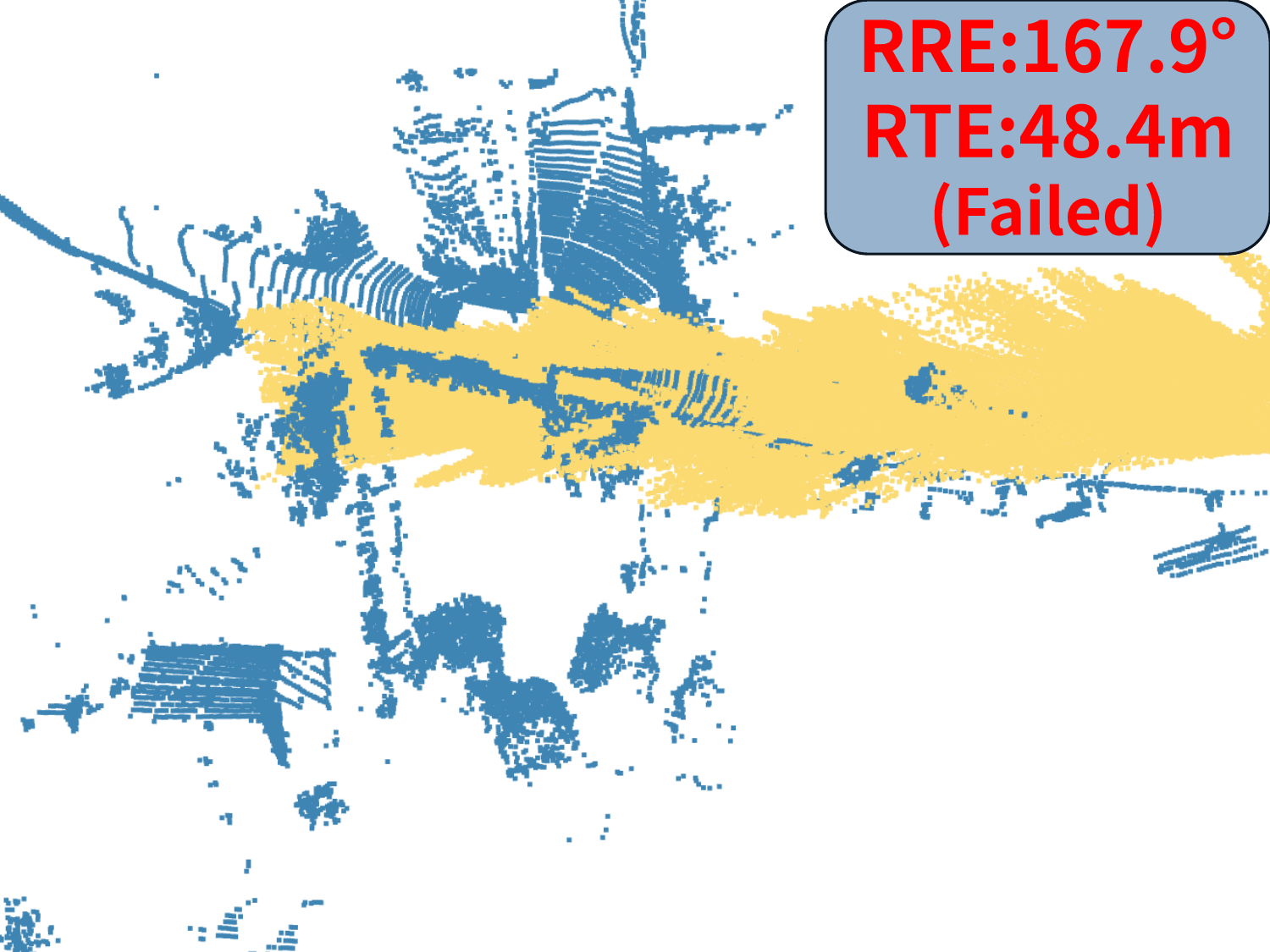}
  }%
  \quad
  \subfigure[hang,center][Superpoint Correspondences (GeoTrans.)]{%
  \label{fig:dcscgeotrans}%
  \includegraphics[width=0.45\linewidth]{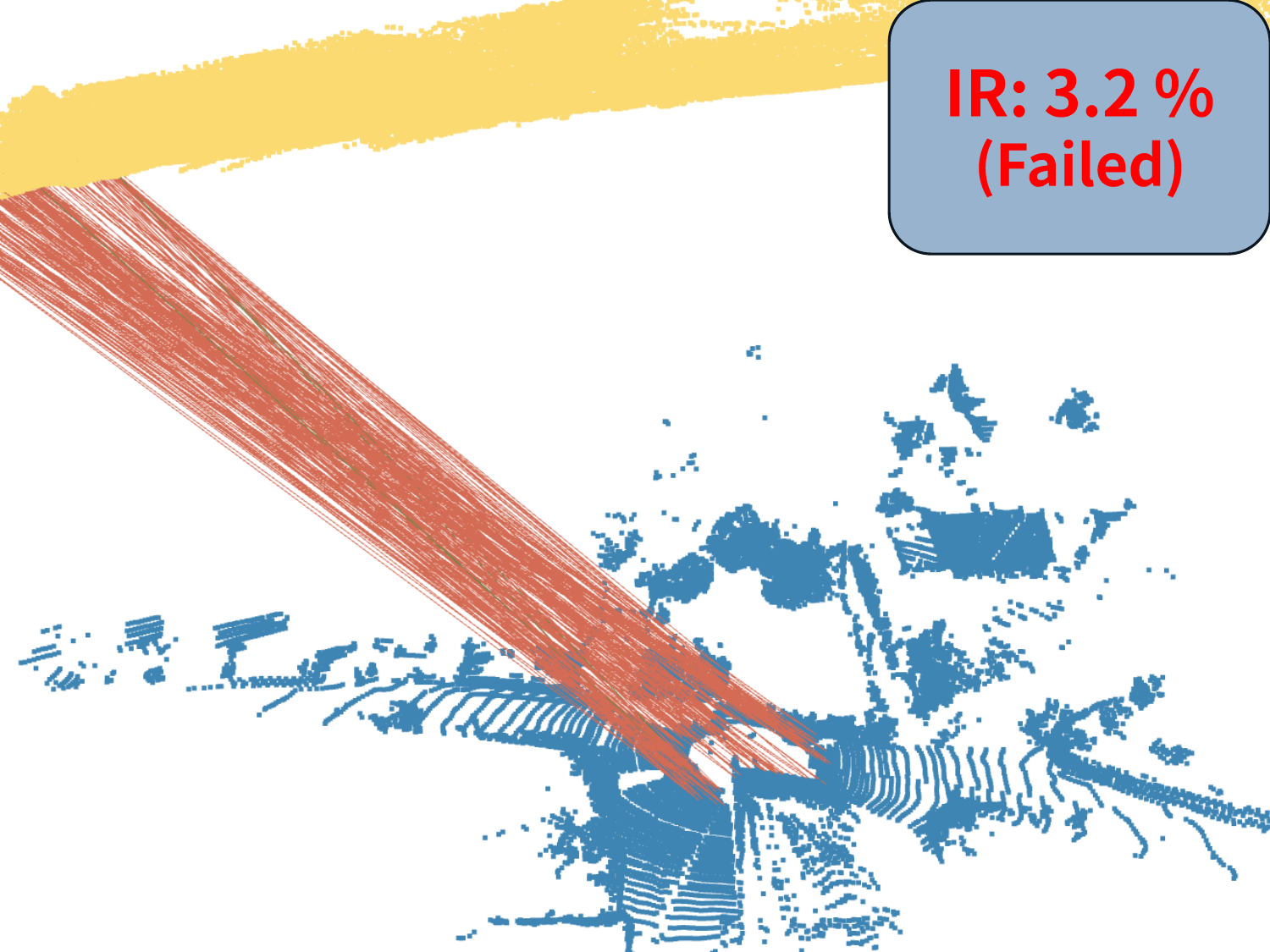}
  }%
  \quad
  \subfigure[Superpoint Correspondences (\mynetwork)]{%
  \label{fig:dcscspeal}%
  \includegraphics[width=0.45\linewidth]{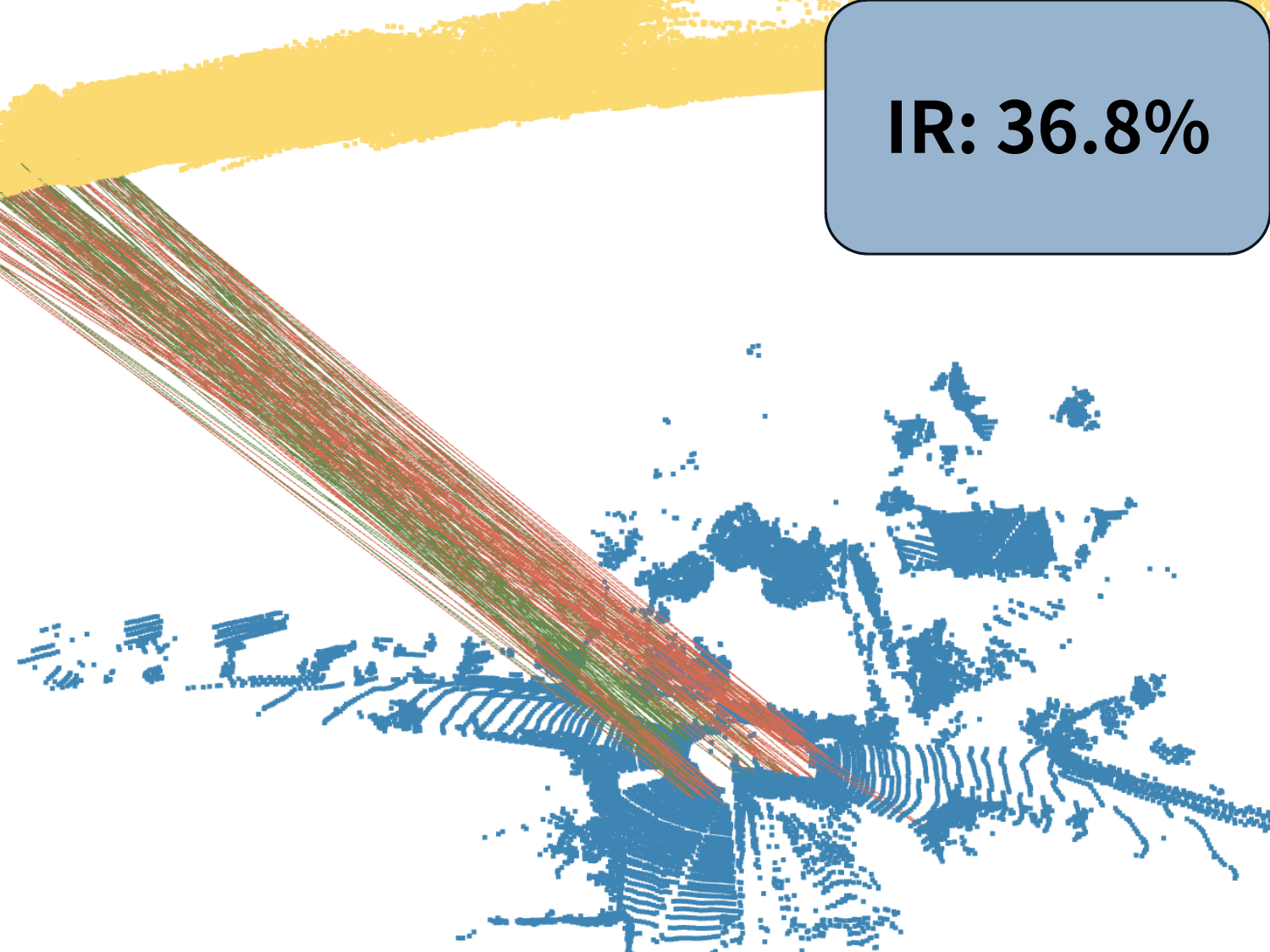}
  }%
  \quad
  \subfigure[Skeletal Correspondences (\mynetwork)]{%
  \label{fig:dcskspeal}%
  \includegraphics[width=0.45\linewidth]{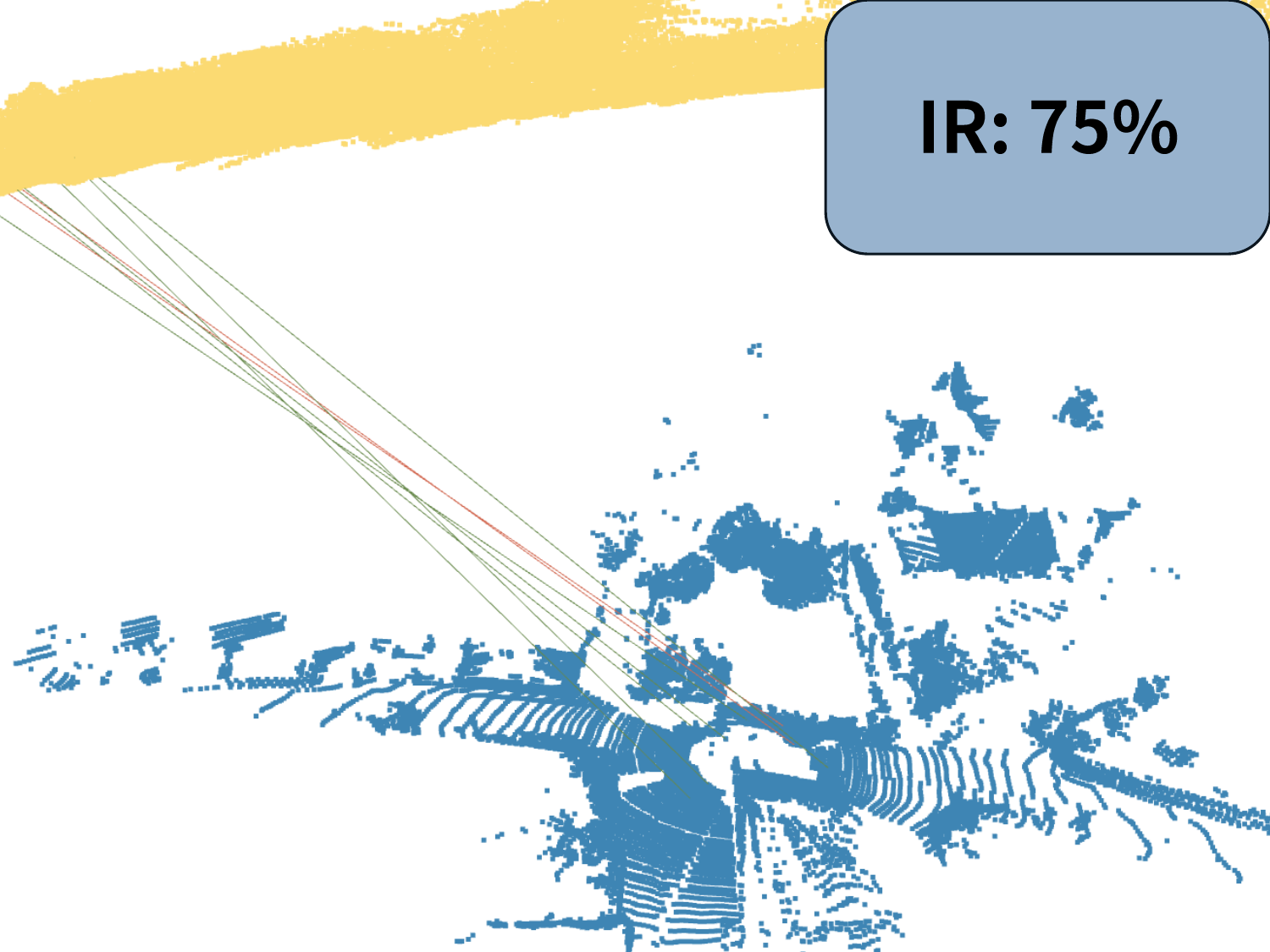}
  }%
  \caption{Qualitative results on \kittics. Red and green denotes outlier and inlier correspondences, respectively.}
  \label{fig:dualcorr}
\end{figure}

\customparagraph{Effectiveness of the Skeletal Priors}
To qualitatively verify the effectiveness of the skeletal representation, we visualize the dual-class correspondences from the \cds{} module, \qs{including superpoints and skeleton points}.
\qs{The qualitative results are shown in Fig.~\ref{fig:dualcorr}.}
In addition to the challenges of partial overlap and density differences, this scan also presents the challenge of unstructured objects.
However, \mynetwork{} is still able to extract right correspondences with the help of skeletons, while the current state-of-the-art, GeoTrans., completely fails.
\mynetwork{} achieves an IR of $36.8\%$, which is nearly $10\times$ higher than GeoTrans. 
It is worth noting that \xkznew{with our spectral denoising step,} the skeletal correspondences achieve an IR of $75\%$, demonstrating the effectiveness of the spectral denoising step. 

\begin{figure}[ht]
  \centering
  \subfigure[Registration recalls with different RRE and RTE thresholds]{%
  \label{fig:threscurves}%
  \includegraphics[width=0.9\linewidth]{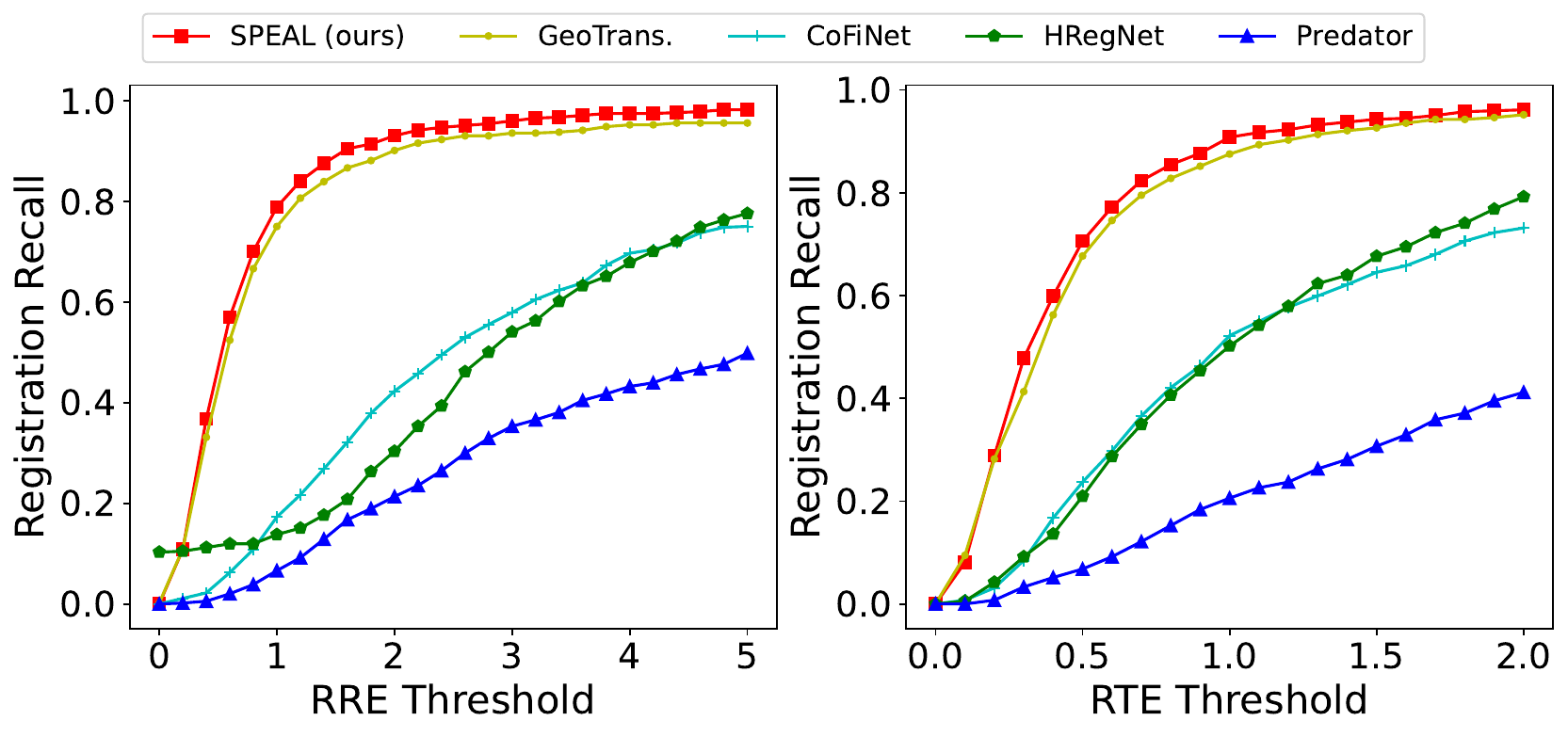}
  }%
  \quad
  \subfigure[Correspondence and registration results under different overlap ratios]{%
  \label{fig:orscurves}%
  \includegraphics[width=0.9\linewidth]{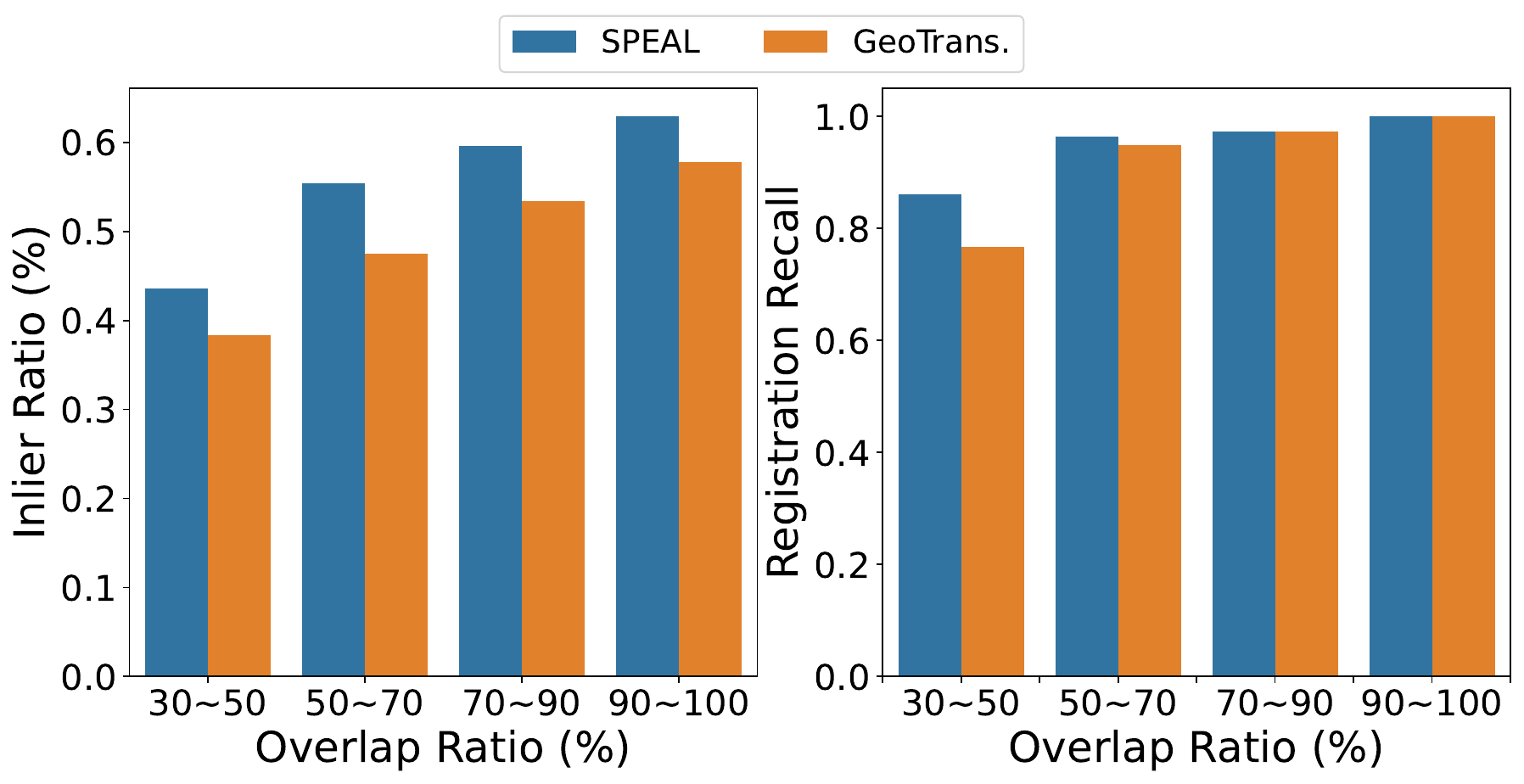}
  }%
  \caption{Quantitative results on robustness of \mynetwork{} on the \kittics{} dataset.}
  \label{fig:robustness}
\end{figure}

\customparagraph{Robustness}
Fig.~\ref{fig:threscurves} displays registration recalls with different RRE and RTE thresholds in \kittics{}.
\mynetwork{} consistently outperforms the other methods. \qs{In particular, it} achieves a registration recall of $86.04\%$ in the challenging low overlap of $30\%\sim50\%$,  which is $9.3\%$ higher than the second-best method.
In addition, Fig.~\ref{fig:orscurves} compares registration recalls and inlier ratios under different overlap with GeoTrans. 
Our method consistently achieves higher inlier ratios under all overlap ratios.
This demonstrates the superior quality of correspondences generated by our method with skeletal priors.
The results prove that our \mynetwork{} is robust to various threshold settings, and demonstrates outstanding performance in low overlap condition.

\subsection{Ablation Studies}

\customparagraph{Overall Effectiveness}
We conduct ablation studies to assess the effectiveness of \mynetwork{} on \germanyforest{}.
We compare different configurations of \mynetwork{}, including: 
(a) vanilla geometric self-attention and vanilla cross attention,
(b) skeleton-aware self-attention and vanilla cross attention,
(c) skeleton-aware self-attention and skeleton-aware cross attention.
In addition, we also compare with 
(d) the method without the \cds{} module, which only samples the coarse correspondences from superpoints.
The results \qs{in Table~\ref{tbl:abls}} demonstrate the effectiveness of our design.

\begin{table}[ht]
  \centering
  \small
  \setlength{\tabcolsep}{6.8pt}
  \begin{tabular}{l|ccc}
    \hline    
    Model & RRE($\degree$)\textdownarrow & RTE($\mathrm{m}$)\textdownarrow & RR($\%$)\textuparrow \\
    \hline
    (a) vanilla &  0.19 & 0.054 & 96.2 \\
    (b) cross attn. w/o SPE & 0.18 & 0.051 & 97.4 \\
    (c) w/ SSE \& SPE  & 0.18 & 0.049 & 97.1 \\
    \hline
    (d) corr. w/o \cds{} & 0.19 & 0.051 & 96.3 \\
    \hline
    (e) \mynetwork{} (Ours) & \textbf{0.16} & \textbf{0.048} & \textbf{99.3} \\
    \hline
  \end{tabular} 
  \caption{Ablation studies on \mynetwork{}.}
  \label{tbl:abls}
\end{table}

\begin{table}[ht]
  \centering
  \small
  \begin{tabular}{cc|ccc}
    \hline    
    \multicolumn{2}{c|}{Spectral Denoising } & \multirow[c]{2}{*}{RRE($\degree$)\textdownarrow} & \multirow[c]{2}{*}{RTE($\mathrm{m}$)\textdownarrow} & \multirow[c]{2}{*}{RR($\%$)\textuparrow} \\
    Sup. Corr. & Skel. Corr. & \\
    \hline
    & & 0.17 & 0.051 & 98.1 \\
    \checkmark & & 0.20 & 0.063 & 97.8\\
    \checkmark & \checkmark & 0.19 & 0.062 & 98.1\\
    \hline
    & \checkmark & \textbf{0.16} & \textbf{0.048} & \textbf{99.3} \\
    \hline
  \end{tabular} 
  \caption{Ablation studies on the \cds{} module. Sup. Corr. and Skel. Corr. denote the superpoint correspondences and skeletal correspondences, respectively.}
  \label{tbl:cdsabl}
\end{table}

\customparagraph{Spectral Denoising in \cds{}}
To validate the effectiveness of spectral denoising, we also \xkznew{ablate} the \cds{} module.
We compare different schemes for applying the spectral denoising.
The results in Table~\ref{tbl:cdsabl} demonstrate that the spectral denoising step is only necessary for the skeletal correspondences and leads to inferior performance in other configurations.

\section{Conclusion}
In this paper, we have proposed \mynetwork, a novel point cloud registration method that leverages a MAT-based skeletal representation to capture the geometric intricacies, thereby \qs{facilitating} registration.
Our method introduces \sem{} to extract the skeleton points and their skeletal features. 
Furthermore, we design \sagtr{} and \cds{} which explicitly integrate skeletal priors to ensure robust and accurate correspondences.
Extensive experiments demonstrate that \mynetwork{} is effective for both same-source and cross-source point cloud registration. 

\customparagraph{Acknowledgement}
This work was supported in part by the National Key R\&D Program of China under Grant 2021YFF0704600, the Fundamental Research Funds for the Central Universities (No. 20720220064).


\bibliography{aaai24}

\end{document}